\documentclass[letterpaper, 10pt, conference]{ieeeconf}  

\IEEEoverridecommandlockouts                              

\usepackage{amsmath} 
\usepackage{amsfonts}
\usepackage{mathtools}

\usepackage{hyperref}

\usepackage{algorithmicx}
\usepackage{algorithm}
\usepackage{algpseudocode}

\usepackage{import}
\usepackage{graphicx}

\usepackage{xcolor}
\usepackage[caption=false]{subfig}
\usepackage[bottom]{footmisc} 

\newcommand{\bm}[1]{\boldsymbol{\mathbf{#1}}}

\usepackage{todonotes}

\newcommand{\trsp}{{^{\top}}}

\newcommand{\etal}{\MakeLowercase{et al.\ }}

\usepackage[nolist]{acronym} 

\newacro{cbf}[CBF]{Control Barrier Function}
\newacro{tvcbf}[TV-CBF]{Time-Varying Control Barrier Function}
\newacro{ccbf}[CCBF]{Conic Control Barrier Function}
\newacro{ccbfqp}[CCBF-QP]{Conic Control Barrier Function Quadratic Programming}
\newacro{tvccbf}[TV-CCBF]{Time-Varying Conic Control Barrier Function}
\newacro{ds}[DS]{Dynamical System}
\newacro{gmm}[GMM]{Gaussian Mixture Model}
\newacro{gmr}[GMR]{Gaussian Mixture Regression}
\newacro{lfd}[LfD]{Learning from Demonstration}
\newacro{dof}[DoF]{Degree of Freedom}
\newacro{em}[EM]{Expectation-Maximization}
\newacro{lwr}[LWR]{Locally Weighted Regression}
\newacro{ee}[EE]{End Effector}
\newacro{qp}[QP]{Quadratic Programming}
\newacro{dmp}[DMP]{Dynamic Motion Primitive}
\newacro{seds}[SEDS]{Stable Estimator of Dynamical Systems}
\newacro{tpgmm}[TP-GMM]{Task-Parameterized Gaussian Mixture Model}
\newacro{physicsgmm}[PC-GMM]{Physically-Consistent Gaussian Mixture Model}
\newacro{nacv}[NACV]{normalized average constraint violation}
\newacro{rlasa}[R-LASA]{Riemannian LASA dataset}
\newacro{}[]{}
\newacro{}[]{}
\newacro{}[]{}
\newacro{}[]{}
\newacro{}[]{}

\title{\LARGE \bf
Safe Execution of Learned Orientation Skills  with Conic Control Barrier Functions
}

\author{Zheng Shen$^{1}$, Matteo Saveriano$^{2}$, Fares J. Abu-Dakka$^{3}$, and Sami Haddadin$^{1}$
\thanks{*This work has been partially supported by the Federal Ministry of Education and Research of Germany in the programme of “Souverän. Digital. Vernetzt.” Joint project 6G-life (16KISK002), the European Union project euROBIN under grant agreement No. 101070596, and the European Union project INVERSE under grant agreement No. 101136067.}
\thanks{$^{1}$Munich Institute of Robotics and Machine Intelligence (MIRMI), Technical University of Munich, Germany.
        }%
\thanks{$^{2}$Department of Industrial Engineering (DII), University of Trento, Trento, Italy.
        }%
\thanks{$^{3}$Electronic and Informatics Department, Faculty of Engineering, Mondragon Unibertsitatea, Bilbao, Spain. 
        }%
\thanks{Corresponding author: {\tt\small zheng.shen@tum.de}}
}
\begin{document}

\maketitle
\thispagestyle{empty}
\pagestyle{empty}

\begin{abstract}
In the field of \ac{lfd}, \acp{ds} have gained significant attention due to their ability to generate real-time motions and reach predefined targets. However, the conventional convergence-centric behavior exhibited by \acp{ds} may fall short in safety-critical tasks, specifically, those requiring precise replication of demonstrated trajectories or strict adherence to constrained regions even in the presence of perturbations or human intervention. Moreover, existing \ac{ds} research often assumes demonstrations solely in Euclidean space, overlooking the crucial aspect of orientation in various applications. To alleviate these shortcomings, we present an innovative approach geared toward ensuring the safe execution of learned orientation skills within constrained regions surrounding a reference trajectory. This involves learning a stable \ac{ds} on $SO(3)$, extracting time-varying conic constraints from the variability observed in expert demonstrations, and bounding the evolution of the \ac{ds} with \ac{ccbf} to fulfill the constraints. We validated our approach through extensive evaluation in simulation and showcased its effectiveness for a cutting skill in the context of assisted teleoperation.

\end{abstract}

\section{INTRODUCTION}

\ac{lfd} enables robots to learn novel skills by imitating human actions instead of coding them~\cite{billard2008survey}. This involves automatic extraction task requirements from human demonstrations. Ideally, the acquired skills are generalizable, agnostic to specific robot platforms, and robust against perturbations. Among \ac{lfd} approaches, \acp{ds} are attractive due to their capability to generate real-time motions and converge toward a predefined target \cite{khansari2011learning}. During the execution phase, a robot's initial state is fed into the \ac{ds}, enabling the robot to navigate and reach its intended goal despite environmental uncertainties or changes. Nevertheless, while this convergence-centric behavior proves effective in various applications, it may be inadequate or even counterproductive in safety-critical tasks that demand the robot to closely mimic the demonstrated trajectories or stay strictly within a constrained region~\cite{figueroa2022locally}. 

\begin{figure}[t!]
\centering
   \def\svgwidth{\columnwidth}
    {\fontsize{8}{8}
\begingroup%
  \makeatletter%
  \providecommand\color[2][]{%
    \errmessage{(Inkscape) Color is used for the text in Inkscape, but the package 'color.sty' is not loaded}%
    \renewcommand\color[2][]{}%
  }%
  \providecommand\transparent[1]{%
    \errmessage{(Inkscape) Transparency is used (non-zero) for the text in Inkscape, but the package 'transparent.sty' is not loaded}%
    \renewcommand\transparent[1]{}%
  }%
  \providecommand\rotatebox[2]{#2}%
  \newcommand*\fsize{\dimexpr\f@size pt\relax}%
  \newcommand*\lineheight[1]{\fontsize{\fsize}{#1\fsize}\selectfont}%
  \ifx\svgwidth\undefined%
    \setlength{\unitlength}{237.47525673bp}%
    \ifx\svgscale\undefined%
      \relax%
    \else%
      \setlength{\unitlength}{\unitlength * \real{\svgscale}}%
    \fi%
  \else%
    \setlength{\unitlength}{\svgwidth}%
  \fi%
  \global\let\svgwidth\undefined%
  \global\let\svgscale\undefined%
  \makeatother%
  \begin{picture}(1,0.67131997)%
    \lineheight{1}%
    \setlength\tabcolsep{0pt}%
    \put(0,0){\includegraphics[width=\unitlength,page=1]{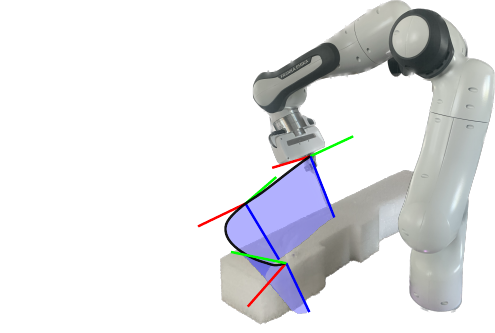}}%
    \put(0.71168522,0.35641631){\color[rgb]{0,0,0}\makebox(0,0)[lt]{\lineheight{1.25}\smash{\begin{tabular}[t]{l}Blade\end{tabular}}}}%
    \put(0.12234501,0.60708481){\color[rgb]{0,0,0}\makebox(0,0)[lt]{\lineheight{1.25}\smash{\begin{tabular}[t]{l}Haptic Device\end{tabular}}}}%
    \put(0.49936105,0.60866338){\color[rgb]{0,0,0}\makebox(0,0)[lt]{\lineheight{1.25}\smash{\begin{tabular}[t]{l}Robot\end{tabular}}}}%
    \put(0.30098071,0.16680161){\color[rgb]{0,0,0}\makebox(0,0)[lt]{\lineheight{1.25}\smash{\begin{tabular}[t]{l}Incision \\Shape\end{tabular}}}}%
    \put(0,0){\includegraphics[width=\unitlength,page=2]{cutting_pose_traj.pdf}}%
  \end{picture}%
\endgroup%
}
     \vspace{-1cm}
\caption{Setup of the experiment for assisted teleoperation. A human operator uses a haptic device to adjust the cutting skill executed by a robot. The $z$-axis of the \ac{ee} frame (blue segment) coincides with the blade. The shaded blue surface indicates the incision shape.}
\label{fig:cutting setup}
\end{figure}

The majority of works on \acp{ds} assume demonstrations in Euclidean space and primarily consider task constraints like obstacle avoidance concerning safety. However, it is clear that orientation, residing in non-Euclidean space, also plays a significant role in various applications. For instance, consider the task of transporting and pouring a cup of water. Here, it becomes crucial to constrain the orientation of the cup to prevent unintended spillage, a requirement that goes beyond the conventional scope of Euclidean task constraints. Similarly, for tissue resection in the domain of robotic surgery, precise orientation of surgical instruments is vital for performing delicate procedures accurately and safely. Indeed, as shown in Fig. \ref{fig:cutting setup}, the orientation trajectory directly influences the shape of the incision.

To address these challenges, we introduce a novel approach that ensures the safe execution of acquired orientation skills when subjected to perturbation or human intervention. This approach guarantees that the execution stays close to a reference trajectory within the region defined by the constraints derived from the variability of demonstrations. The primary contributions of this work are: (\emph{i}) An extension of \ac{physicsgmm}~\cite{figueroa2018physically} for learning stable \acp{ds} on $SO(3)$. Notably, we provide a formal proof of its asymptotic stability.  (\emph{ii}) A novel approach that extracts orientation constraints based on the variability of demonstrations and exploits \acf{ccbf}~\cite{ibuki2020distributed} to ensure safe execution.  (\emph{iii}) Validation of our method in simulation and in an assisted teleoperation experiment focused on robotic cutting tasks. 
\section{Related Work}

A significant body of \ac{lfd} literature focuses on stable motion execution. One of the most prevalent methods, the \ac{dmp}~\cite{ijspeert2013dynamical,saveriano2021dynamic}, employs a phase variable to cancel the nonlinear forcing term superimposed on a stable linear \ac{ds} to ensure convergence. \ac{dmp}, being a deterministic method, is limited to learning from a single demonstration. The \ac{seds}~\cite{khansari2011learning} approximates a nonlinear \ac{ds} with a combination of linear ones but faces accuracy and stability dilemmas. 
\ac{physicsgmm}~\cite{figueroa2018physically} alleviates this limitation by decoupling the parameters of \ac{gmm} from those of the linear \acp{ds}. The \ac{ds} formulation in~\cite{figueroa2022locally} provides a behavior akin to ``trajectory-tracking'' for specific reference trajectories and retains global convergence. Although the aforementioned approaches are effective in learning position trajectories, they neglect orientation. In \cite{zeestraten2017approach}, Riemannian metrics are used to learn orientation trajectories via task-parameterized \ac{gmm}, without considering stability. Abu-Dakka~\etal\cite{abu2022unified} extended \ac{dmp} to Riemannian manifolds like quaternions, while Saveriano~\etal\cite{saveriano2023learning} learnd a diffeomorphism to map simple manifold trajectories into complex ones. In our current work, we draw inspiration from the superior performance of \ac{physicsgmm} \cite{figueroa2018physically} and extend the framework to orientation and establish its asymptotic stability on $SO(3)$.

Learning constraints from demonstrations have been explored in the literature. In~\cite{ureche2015task}, position and force constraints are extracted from demonstration variance. Menner~\etal\cite{menner2019constrained} learned cost functions and constraints simultaneously in an inverse optimal control framework, but the constraint is limited to a convex hull in Euclidean space. In \cite{saveriano2019learning}, linear subspaces are learned as constraints incrementally and used \acp{cbf} to generate constrained motions. \acp{cbf} were also used in~\cite{davoodi2021probabilistic} to maintain a joint space trajectory within a learned covariance bound. A neural network~\cite{sutanto2021learning} and kernelized principle component analysis~\cite{mehr2016inferring} are used to learn equality constraints as manifolds, whereas the learned constraints are on feature space and lack of explainability. Perez-D’Arpino~\etal\cite{perez2017c} utilized task space region, a volume in $SE(3)$ around the keyframes, for constrained motion planning. Chou~\etal\cite{chou2021learning} proposed a method to learn grid and parametric constraints from demonstrations and uses Euler angles to represent the orientation. Like most works, we exploit the variability in demonstrations and extract orientation constraints based on conic constraints. 

 \acp{cbf} render a set forward invariant and are thus widely used in the control of constrained systems~\cite{ames2019control}. \ac{tvcbf}~\cite{IGARASHI2019735} can be used to enforce time-varying constraints. Wu and Sreenath~\cite{wu2016safety} extended \acp{cbf} from Euclidean space to manifolds with applications to geometrical control of mechanical systems under time-varying constraints. In~\cite{ibuki2020distributed}, \acp{ccbf} are used to solve a distributed collision avoidance problem for a group of agents on a sphere. Tan~\etal\cite{tan2020construction} proposed a union of hyperspherical constraints as \ac{cbf} on $SO(3)$ and applied it to safety-critical control along a $\mathcal{C}^2$ reference trajectory. In comparison, a conic constraint is preferred in this work as it provides more flexibility when constructing constraints. 

In assisted teleoperation, robots aid in task completion and simplify teleoperation for the human operator. The most prevalent method is virtual fixture~\cite{rosenberg1993virtual}, which assists the operator by guiding or constraining either the user's input or the robot's motion. Ewerton~\etal\cite{ewerton2020assisted} provided haptic cues for goal-reaching tasks by constructing a potential field based on learned \ac{gmm} over demonstrations. Dragan and Srinivasa~\cite{dragan2013policy} formalized assisted teleoperation as a policy blending problem, arbitrating human input and robot action based on the confidence of prediction of human intention. The prediction of intended goals was further improved by solving a partially observable Markov decision process with hindsight optimization~\cite{javdani2018shared}. \ac{tpgmm} encodes demonstrated trajectories and blends the learned model and user inputs with a linear quadratic regulator \cite{havoutis2016learning}. In \cite{havoutis2017supervisory}, the remote robot autonomously executed a task learned with a task-parameterized hidden \ac{gmm}, with the user only providing high-level task goals. Mower~\etal\cite{mower2021skill} proposed a receding horizon shared control method, which adapts future trajectories based on the estimation of the operator's intended skill. We show the applicability of our method to assisted teleoperation while providing formal safety guarantees.

\section{Preliminaries}
\subsection{Special Orthogonal Group ${SO(3)}$}\label{subsec:so3}

Traditionally, orientations in 3D-space are represented as rotation matrices that inhabit the Special Orthogonal Group denoted as ${SO(3)}$~\cite{bullo2019geometric}, where
\begin{equation*}
    SO(3):=\left\{\bm{R}\in \mathbb{R}^{3 \times 3} \mid \bm{R}\trsp \bm{R}=\bm{R} \bm{R}\trsp=\bm{I},\, \operatorname{det}(\bm{R})=1\}\right. .
\end{equation*}
The group equipped with matrix multiplication constitutes a Lie Group. The associated Lie algebra comprises all $3\times 3$ skew-symmetric matrices, i.e., ${\mathfrak{s o}(3):=\left\{\bm{\Omega} \in \mathbb{R}^{3 \times 3} \mid \bm{\Omega}\trsp = - \bm{\Omega}\}\right.}$. The mapping $(\cdot)^{\wedge}$, also denoted as ${[(\cdot)]_{\times}:\mathbb{R}^3 \rightarrow \mathfrak{s o}(3)}$ , along with its inverse mapping ${(\cdot)^{\vee}: \mathfrak{s o}(3) \rightarrow \mathbb{R}^3}$ are defined as 
\begin{equation*}
    \bm{x}=\left(\begin{smallmatrix}x_1 \\ x_2 \\ x_3\end{smallmatrix}\right) \stackrel{[(\cdot)]_{\times}}{\underset{(\cdot)^{\vee}}{\rightleftharpoons}} [\bm{x}]_{\times} = \left(\begin{smallmatrix}0 & -x_3 & x_2 \\ x_3 & 0 & -x_1 \\ -x_2 & x_1 & 0\end{smallmatrix}\right).
\end{equation*}
Throughout the paper, we refer to $\bm{x}$ such that $[\bm{x}]_{\times} \in \mathfrak{s o}(3)$ as a tangent vector. 

For a given $[\bm{x}]_{\times} \in \mathfrak{s o}(3)$, the exponential map function ${\operatorname{exp}(\cdot): \mathfrak{s o}(3) \rightarrow SO(3)}$ allows the representation of $[\bm{x}]_{\times}$ as rotation matrices
\begin{equation}
    \exp \left([\bm{x}]_{\times}\right)= 
    \begin{cases}
        {\bm{I}+\frac{\sin \left(\|\bm{x}\|\right)}{\|\bm{x}\|}[\bm{x}]_{\times}+\frac{1-\cos\left(\|\bm{x}\|\right)}{\|\bm{x}\|^2}[\bm{x}]_{\times}^2}, \ \bm{x} \neq 0 \\
        \bm{I}, \text { otherwise. }
    \end{cases}
\end{equation}
Its inverse is the logarithmic map $\operatorname{log}(\cdot): SO(3) \rightarrow \mathfrak{s o}(3) $
\begin{equation}
\operatorname{log}(\bm{R}) = 
\begin{cases}
    \frac{\theta(\bm{R})}{2 \sin (\theta(\bm{R}))}\left(\bm{R}-\bm{R}\trsp\right), \ \bm{R} \neq \bm{I} \\
    \bm{0}, \ \text{otherwise,}
\end{cases}
\end{equation}
where $\theta({\bm{R}}) := \arccos ((\operatorname{tr}(\bm{R})-1) / 2)$. To simplify the notation, we introduce the uppercase exponential map as $\operatorname{Exp}(\bm{x}):=\exp \left([\bm{x}]_{\times}\right)$and the uppercase logarithmic map as ${\operatorname{Log}\left(\bm{R}\right) := \left(\operatorname{log}\left(\bm{R}\right)\right)^{\vee}}$. Additionally, we define $\operatorname{Log}_{\bm{R}_b}(\bm{R}_a):= \operatorname{Log}(\bm{R}_a^{\top}\bm{R}_b)$.


\subsection{Time-Varying Control Barrier Function}
 
Consider the control-affine system
\begin{equation}\label{eq:affine-control DS}
\dot{\bm{x}}=\bm{f}(\bm{x}) + \bm{g}(\bm{x})\bm{u},
\end{equation}
where $\mathbf{x} \in \mathcal{X} \subset \mathbb{R}^n$ is the state, $\mathbf{u} \in \mathcal{U} \subset \mathbb{R}^m$ is the control input, and $\mathbf{f}: \mathbb{R}^n \rightarrow \mathbb{R}^n$  and $\mathbf{g}: \mathbb{R}^n \rightarrow \mathbb{R}^{n \times m}$ are locally Lipschitz continuous functions.

\textbf{Definition 1} (Extended class $\mathcal{K}_{\infty}$ function): A continuous function $\alpha:(-b, a) \rightarrow \mathbb{R}$ belongs to the extended class $\mathcal{K}_{\infty}$ for some $a, b>0$, if $\alpha(0)=0$ and $\alpha$ increases strictly monotonically.

\textbf{Definition 2} (forward invariance of time-varying set)  
A time-varying set $\mathcal{C}(t) \subset \mathbb{R}^n$ is forward invariant to \eqref{eq:affine-control DS} for a given control law $\bm{u}$, if for any $\bm{x}_0 \in \mathcal{C}(t_0)$, there exists a unique solution $\bm{\phi}: [t_0, t_1] \rightarrow \mathbb{R}^n$ with $\bm{\phi}(t_0) = \bm{x}_0$ and $\frac{d}{d t}\bm{\phi}(t) = \bm{f}(\bm{\phi}(t)) + \bm{g}(\bm{\phi}(t))\bm{u}$ such that $\bm{\phi}(t) \in \mathcal{C}(t)$ for all $t \in [t_0, t_1]$.

\textbf{Definition 3} (time-varying control barrier function): Let $\mathcal{C}(t) :=\left\{\bm{x} \in \mathcal{X} \subset \mathbb{R}^n \mid h(\bm{x}, t) \geq 0\right\}$ be the 0-superlevel set of a smooth function $h(\bm{x}, t): \mathbb{R}^n \times [t_0, t_1] \rightarrow \mathbb{R}$, then $ h $ is a \ac{tvcbf} if there exist an extended class $\mathcal{K}_{\infty}$ function $\alpha$ such that for all $(\bm{x}, t) \in \mathcal{X} \times [t_0, t_1]$, the control system in \eqref{eq:affine-control DS} satisfies
\begin{equation}\label{eq:tvCBF condition}
    \sup _{\bm{u} \in \mathcal{U}}\left[L_{\bm{f}} h(\bm{x}, t)+L_{\bm{g}} h(\bm{x}, t) \bm{u}\right] + \frac{\partial h(\bm{x}, t)}{\partial t} \geq-\alpha(h(\bm{x}, t)).
\end{equation}
Applying any Lipschitz continuous controller $u(\bm{x}, t) \in \mathcal{U}(\mathbf{x}, t)$ to \eqref{eq:affine-control DS}, $\mathcal{U}(\bm{x}, t)=\{\bm{u} \in \mathcal{U} \mid L_{\bm{f}} h(\bm{x}, t)+L_{\bm{g}} h(\bm{x}, t) \bm{u}+ \frac{\partial h(\bm{x}, t)}{\partial t} +\alpha(h(\bm{x}, t)) \geq 0\}$, the forward invariance of $\mathcal{C}(t)$ is guaranteed \cite[Theorem 1] {lindemann2018control}, which enables us to use the following \ac{qp} that minimally modifies the reference controller $\bm{u}_0(\bm{x}, t)$:
\begin{equation}\label{eq:CBF QP}
\bm{u}^*(\bm{x}, t)=\underset{\bm{u} \in \mathcal{U}}{\operatorname{argmin}} \frac{1}{2}\|\bm{u}-\bm{u}_0(\bm{x}, t)\|_2^2 \text { subject to Eq. \eqref{eq:tvCBF condition}}.
\end{equation}

\subsection{Conic Control Barrier Function} 
Similarly to the case of Euclidean space, we can define a \acp{cbf} on $ SO(3)$ as follows:

\textbf{Definition 4} (\ac{cbf} on $SO(3)$): Consider the 0-superlevel set $\mathcal{C}=\left\{\bm{R} \in \mathcal{S} \subset SO(3) \mid h(\bm{R}) \geq 0\right\}$ of a smooth function $h(\bm{R}): SO(3) \rightarrow \mathbb{R}$, then $ h$ is a \ac{cbf} on $SO(3)$ if there exists an extended class $\mathcal{K}_{\infty}$ function $\alpha$ such that for all $\bm{R} \in \mathcal{S} $ the attitude dynamics $\dot{\bm{R}} = \bm{R}[\bm{\omega}]_{\times}$ satisfies
\begin{equation}
\sup _{\bm{\omega} \in \bm{\Omega}}\left(\dot{h}(\bm{R}) \right) \geq-\alpha\left(h(\bm{R})\right)
\end{equation}

\begin{figure}[t!]
\centering
    \def\svgwidth{0.5\columnwidth}
    {\fontsize{9}{9}
    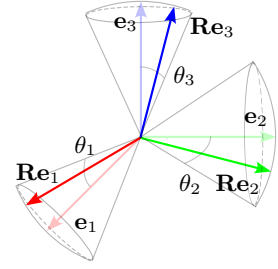}
\caption{An illustration of conic constraints.}
\label{fig:conic_constraint}
\end{figure}

One option for $h(\bm{R})$ is conic attitude constraints. Consider the inequality constraint as follows:
\begin{equation}
h_i(\bm{R}) := \bm{e}_i\trsp\bm{R} \bm{e}_i - \cos(\theta_i) \geq 0,
\end{equation}
where $\theta_i \in(0, \pi / 2)$ determines the size of the cone and $\bm{e}_i = [\delta_{i1}\ \delta_{i2}\ \delta_{i3}]^{\top}$, with $\delta_{ij}$ being the Kronecker delta and $i, j \in \{1, 2, 3\}$.
As illustrated in Fig.\ref{fig:conic_constraint}, the transformed axis $\bm{R}\bm{e}_i$ is constrained in the conic region determined by $\bm{e}_i$ and $\theta_i$. Ibuki~\etal\cite{ibuki2020distributed} proves that for the attitude dynamics,  any Lipschitz continuous control law $\bm{\omega}(\bm{R}) \in \bm{\Omega}(\bm{R})$, $\bm{\Omega}(\bm{R})=\left\{\bm{\omega} \in \bm{\Omega} \mid -\bm{e}_i\trsp\bm{R}[\bm{e}_i]_{\times}\bm{\omega} + \alpha(\bm{e}_i\trsp\bm{R} \bm{e}_i - \cos(\theta_i)) \geq 0\right\}$ will render the set $\mathcal{C}$ forward invariant\footnote{The property $[\bm{a}]_{\times}\bm{b} = - [\bm{b}]_{\times}\bm{a}$ is used to obtain the \ac{ccbf} condition.}.

\section{METHODOLOGY}

\begin{figure}[b!]
\centering
    \def\svgwidth{\columnwidth}
    {\fontsize{9}{9}
    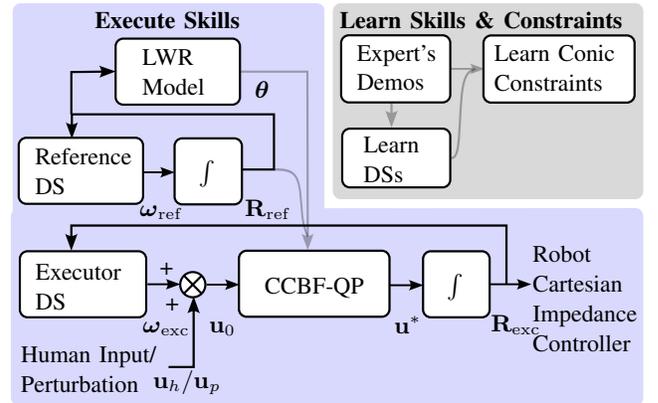}
\caption{Overview of the method. Dark-colored lines indicate values. Light-colored lines indicate data or functions.}
\label{fig:overview}
\end{figure}

Our method comprises two phases: an offline phase dedicated to skill and constraint acquisition from demonstrations, and an online phase focused on the safe execution of these skills within the regions defined by the constraints. Fig. \ref{fig:overview} provides an overview of the method. 

During the learning phase, \ac{physicsgmm} encodes translational and rotational motions as two stable \acp{ds}, extracting time-varying conic constraints for orientation based on observed demonstration variability. Specifically, the cone axes align with the learned \ac{ds}, with cone angles $\bm{\theta}$ representing orientation variances encoded in a \ac{lwr} model. In the execution phase, each iteration adjusts the command $\bm{\omega}_\mathrm{exc}$ from the learned \ac{ds} with human input $\bm{u}_h$ or perturbed by noise $\bm{u}_p$. The summation $\bm{u}_0$ serves as the reference control input for the \ac{ccbfqp}, which acts as a filter that minimally modifies the given input while guaranteeing forward invariance of the safety set defined by the time-varying conic constraints. Finally, we obtain the trajectory to be executed $\bm{R}_\mathrm{exc}(t)$ by discretely integrating the solution $\bm{u}^*$ to the \ac{qp}.

\subsection{Learning from Demonstration}\label{subsection:lfd}
We formulate robot motions $ (\bm{p}, \bm{R}) \in \mathbb{R}^3\times SO(3)$ as a control law driven by two separate \acp{ds} that encode a specific behavior. From a machine learning perspective, estimating the \acp{ds} from a set of $N$ reference trajectories $\mathcal{D} = \left\{\left(\bm{p}^{t, n}, \bm{R}^{t, n}\right), \left(\dot{\bm{p}}^{t, n}, \bm{\omega}^{t, n}\right)\right\}_{t=0, n=1}^{T^n, N}$ can be framed as a regression problem. In this work, we use \ac{physicsgmm} to learn the \acp{ds}~\cite{figueroa2018physically}. For translation, we use a first-order \ac{ds}
\begin{equation}
\dot{\bm{p}} = \bm{f}_{\bm{p}}(\bm{p}),\  \bm{f}_{\bm{p}}: \mathbb{R}^3 \rightarrow \mathbb{R}^3.
\end{equation}

For $\bm{R}\in SO(3)$, given the aim of modeling robots' motion, we assume they follow the rigid body motion \cite{murray2017mathematical}
\begin{equation}\label{eq:rbm}
    \dot{\bm{R}} = \bm{R}[\bm{\omega}]_{\times}.
\end{equation}
The capitalized logarithmic map allows the regression of rotation trajectories to stay in $\mathbb{R}^3$, with tangent vector as inputs and angular velocity $\bm{\omega}$ as outputs
\begin{equation}\label{eq:so3ds}
    {\bm{\omega}} = \bm{f}_{\bm{R}}\left(\operatorname{Log}_{\bm{R}_g}\left(\bm{R}\right)\right), \ \bm{f}_{\bm{R}}: \mathbb{R}^{3} \rightarrow \mathbb{R}^{3},
\end{equation}
where $\bm{R}_g$ is the rotational goal.

Inspired by \ac{physicsgmm}~\cite{figueroa2018physically}, we define the regression function in~\eqref{eq:so3ds} as
\begin{equation}
\label{eq:gmr_rotation}
\bm{\omega} = \sum_{k=1}^K \bm{A}_k \operatorname{Log}_{\bm{R}_g}(\bm{R}),
\end{equation}
where $\bm{A}_k$ are positive definite matrices learned from demonstrations.
In order to prove the stability of~\eqref{eq:gmr_rotation}, we consider the trace Lyapunov candidate $
V(\bm{R}) = \mathrm{tr}(\bm{I} - \bm{R}^{\top}\bm{R}_g)$, that is widely used in attitude control on $SO(3)$~\cite{Berkane2017Construction}. The time derivative of $V(\bm{R})$ writes as:
\begin{equation}
    \label{eq:lyapunov_derivative}
    \dot{V}(\bm{R}) = - \mathrm{tr}(\dot{\bm{R}}^{\top}\bm{R}_g) = - \mathrm{tr}([\bm{\omega}]_{\times}^{\top} \bm{R}^{\top}\bm{R}_g),
\end{equation}
where we used the definition of $\dot{\bm{R}}$ in~\eqref{eq:rbm}. Defining $\bm{E} = \bm{R}^{\top}\bm{R}_g$ and recalling the identity $\mathrm{tr}([\bm{\omega}]_{\times}^{\top} \bm{E})=\bm{\omega}^{\top}\left[(\bm{E} - \bm{E}^{\top}) \right]^{\vee}$~\cite{Berkane2017Construction}, the derivative of the Lyapunov function in~\eqref{eq:lyapunov_derivative} becomes
\begin{equation}
    \label{eq:lyapunov_derivative_2}
    \begin{split}
    \dot{V}(\bm{R}) &= - \bm{\omega}^{\top} \left[(\bm{E} - \bm{E}^{\top}) \right]^{\vee} \\ &= - \sum_{k=1}^K  \left( \operatorname{Log}_{\bm{R}_g}(\bm{R}) \right)^{\top} \bm{A}_k \left[(\bm{E} - \bm{E}^{\top}) \right]^{\vee} ,
    \end{split}
\end{equation}
where we used the definition of $\bm{\omega}$ in~\eqref{eq:gmr_rotation} and the property $\bm{A}_k^{\top}=\bm{A}_k$. From the definition of $\operatorname{Log}_{\bm{R}_g}(\bm{R})$ in Sec.~\ref{subsec:so3}, it is easy to verify that $\dot{V}(\bm{R})$ vanishes at the equilibrium, i.e., $\dot{V}(\bm{R}) = 0$ for $\bm{R} = \bm{R}_g$. For $\bm{R} \neq \bm{R}_g$, i.e., for $\bm{E} \neq \bm{I}$, it holds that
\begin{equation}
\label{eq:log_projection}
    \operatorname{Log}_{\bm{R}_g}(\bm{R}) = \frac{\theta(\bm{E})}{2 \sin (\theta(\bm{E}))}\left[(\bm{E} - \bm{E}^{\top}) \right]^{\vee}.
\end{equation}
By substituting~\eqref{eq:log_projection} into~\eqref{eq:lyapunov_derivative_2}, we obtain that 
\begin{equation}
    \small
    \label{eq:lyapunov_derivative_3}
    \dot{V}(\bm{R}) = - \sum_{k=1}^K \frac{\theta(\bm{E})}{2 \sin (\theta(\bm{E}))}\left(\left[(\bm{E} - \bm{E}^{\top}) \right]^{\vee}\right)^{\top} \bm{A}_k \left[(\bm{E} - \bm{E}^{\top}) \right]^{\vee}.
\end{equation}
Recalling that $\bm{A}_k$ are positive definite matrices and that $\frac{\theta(\bm{E})}{\sin (\theta(\bm{E}))}\geq 0$ for $-\pi < \theta(\bm{E}) < \pi$ and it vanishes only at $\theta(\bm{E}) = 0$ (corresponding to $\bm{E} = \bm{I}$), we conclude the asymptotic stability of~\eqref{eq:gmr_rotation}.

\subsection{Learn time-varying conic constraints}
\begin{algorithm}[b]
\caption{\sc{Learn Cone Angle}}\label{alg:angle}
\begin{algorithmic}[1]
\Require $\left\{\bm{R}^{t, n}\right\}_{t=0, n=1}^{T^n, N}$
\State $D_{min} = \operatorname{min}_{n=1}^{N}\|\operatorname{Log}_{\bm{R}_g^n}(\bm{R}^{0, n})\|_2$
\State $d^i = (D_{\text{min}} / M)\cdot i \textbf{ for }  i = 1:M$
\State $ \left\{\bm{R}^{i, n}\right\}_{i=0, n=1}^{M, N} \gets \texttt{Resample}(\texttt{Slerp}(\left\{\bm{R}^{t, n}\right\}_{t=0, n=1}^{T^n, N})) $
\For{$i \leq M $}
\State $ \bm{m} \gets \texttt{mean}(\left\{\operatorname{Log}_{\bm{R}_g}\left(\bm{R}^{i, n}\right)\right\}_{n=1}^{N})$
\State $ \bm{V} \gets \texttt{covariance}(\left\{\operatorname{Log}_{\bm{R}_g}\left(\bm{R}^{i, n}\right)\right\}_{n=1}^{N})$
\State $ \bm{v}_{1:3} \gets \texttt{Eigen\_Decomposition}(\bm{V})$
\State $ \bm{p}_{1:3} \gets \bm{v}_{1:3} + \bm{m}$
\State $\bm{R}_{1:3}, \bm{R}_m \gets \operatorname{Exp}(\bm{p}_{1:3}), \operatorname{Exp}(\bm{m}) $
\State $\theta_{1:3}^i \gets \operatorname{max}_{j \in \{1,2,3\}}\operatorname{arccos}(\bm{e}_{1:3}^\mathrm{w}\trsp\bm{R}_{j}\trsp\bm{R}_m\bm{e}_{1:3}^\mathrm{w})$
\EndFor
\State $\bm{f}_{\theta}\gets \texttt{LWR}(\{d^i, \theta_{1:3}^i\}_{i=0}^M)$

\State \Return $ \bm{f}_{\theta} $
\end{algorithmic}
\end{algorithm}
We derive time-varying conic constraints, denoted as $h(\bm{R}, t)$, by independently learning cone axes and cone angles. We first outline the definition of the conic axes. By integrating Eq. \eqref{eq:rbm} and Eq. \eqref{eq:so3ds} with a given initial point and goal, we reproduce the reference rotational trajectory $\bm{R}_\mathrm{ref}(t)$. Referring to the axes of the world frame as $\bm{e}_{1:3}^{\mathrm{w}}$, the cone axes are straightforwardly established as the axes of the transformed world frame, denoted as $\bm{e}_{1:3}^\mathrm{ref}(t):=\bm{R}_\mathrm{ref}(t)\bm{e}_{1:3}^{\mathrm{w}}$.

Algorithm \ref{alg:angle} outlines the procedure to encode the variance of orientations as cone angles along demonstrated trajectories by learning the \ac{lwr} model $[\theta_1, \theta_2, \theta_3]\trsp=\bm{f}_{\theta}(\operatorname{Log}_{\bm{R}_g}(\bm{R}))$. It begins by computing the minimum distance \(D_{min}\) between the initial and goal rotations across all demonstrations. This distance is then uniformly partitioned to produce the dataset \(\{d_i\}_{i=0}^M\). Subsequently, we resample the time-dependent demonstrations \(\left\{\bm{R}^{t, n}\right\}_{t=0, n=1}^{T^n, N}\) w.r.t. \(\{d_i\}_{i=0}^M\). For each set of rotational matrices at the same distance, we calculate the mean and covariance of their tangent vectors. Projecting the sum of the mean and the covariance eigenvectors back onto $SO(3)$ yields the mean rotational matrix \(\bm{R}_m\) and three additional rotational matrices \(\bm{R}_{1:3}\). The cone angle for each axis \(\bm{e}_i\) of a reference frame is defined as the maximum cosine angle between the transformed axes \(\bm{R}_m\bm{e}_i\) and \(\bm{R}_j\bm{e}_i\). Finally, we apply \ac{lwr} to the dataset \(\{d^i, \theta_{1:3}^i\}_{i=0}^M \) and obtain the regression model $\bm{f}(\bm{\theta})$.
In conclusion, we establish the time-varying conic constraints as:
\begin{equation}\label{eq:learncc}
h_i(\bm{R}, t) := \bm{e}_i^{\mathrm{w}}\trsp\bm{R}\trsp\bm{e}_i^\mathrm{ref}(t) - \operatorname{cos}(\theta_i(t)) \geq 0, i \in\{1,2,3\},
\end{equation}
with $\theta_i(t) = {\bm{f}_{\theta}}_i(\|\operatorname{Log}_{\bm{R}_g}(\bm{R}_\mathrm{ref}(t))\|_2)$.

\subsection{Constrained execution via \texorpdfstring{\ac{ccbfqp}}{}}
\begin{figure}[t!]
\centering
    \def\svgwidth{0.5\columnwidth}
    {\fontsize{9}{9}
\begingroup%
  \makeatletter%
  \providecommand\color[2][]{%
    \errmessage{(Inkscape) Color is used for the text in Inkscape, but the package 'color.sty' is not loaded}%
    \renewcommand\color[2][]{}%
  }%
  \providecommand\transparent[1]{%
    \errmessage{(Inkscape) Transparency is used (non-zero) for the text in Inkscape, but the package 'transparent.sty' is not loaded}%
    \renewcommand\transparent[1]{}%
  }%
  \providecommand\rotatebox[2]{#2}%
  \newcommand*\fsize{\dimexpr\f@size pt\relax}%
  \newcommand*\lineheight[1]{\fontsize{\fsize}{#1\fsize}\selectfont}%
  \ifx\svgwidth\undefined%
    \setlength{\unitlength}{176.96664885bp}%
    \ifx\svgscale\undefined%
      \relax%
    \else%
      \setlength{\unitlength}{\unitlength * \real{\svgscale}}%
    \fi%
  \else%
    \setlength{\unitlength}{\svgwidth}%
  \fi%
  \global\let\svgwidth\undefined%
  \global\let\svgscale\undefined%
  \makeatother%
  \begin{picture}(1,0.7330967)%
    \lineheight{1}%
    \setlength\tabcolsep{0pt}%
    \put(0,0){\includegraphics[width=\unitlength,page=1]{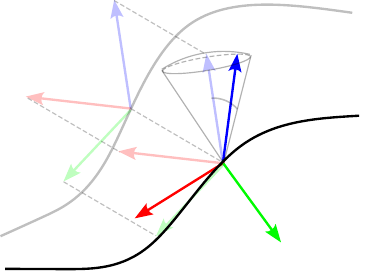}}%
    \put(0.06078179,0.37925717){\color[rgb]{0,0,0}\makebox(0,0)[lt]{\lineheight{1.25}\smash{\begin{tabular}[t]{l}$\bm{p}_\mathrm{ref}(t), \bm{R}_\mathrm{ref}(t)$\end{tabular}}}}%
    \put(0.64294974,0.27512977){\color[rgb]{0,0,0}\makebox(0,0)[lt]{\lineheight{1.25}\smash{\begin{tabular}[t]{l}$\bm{p}_\mathrm{exc}(t), \bm{R}_\mathrm{exc}(t)$\end{tabular}}}}%
    \put(0.66308201,0.41040396){\color[rgb]{0,0,0}\makebox(0,0)[lt]{\lineheight{1.25}\smash{\begin{tabular}[t]{l}$\theta_3(t)$\end{tabular}}}}%
    \put(0.34153403,0.67417127){\color[rgb]{0,0,0}\makebox(0,0)[lt]{\lineheight{1.25}\smash{\begin{tabular}[t]{l}$\bm{e}_3^\mathrm{ref}(t)$\end{tabular}}}}%
    \put(0.67416051,0.53486206){\color[rgb]{0,0,0}\makebox(0,0)[lt]{\lineheight{1.25}\smash{\begin{tabular}[t]{l}$\bm{e}_3^\mathrm{exc}(t)$\end{tabular}}}}%
  \end{picture}%
\endgroup%
}
\caption{An illustration of time-varying conic constraints. The axes of a rotational matrix $\bm{R}_\mathrm{exc}$ always stay within the cones defined by $\bm{R}_\mathrm{ref}(t)$, which itself evolves according to the reference \ac{ds}.}
\label{fig:CCBF_illustration}
\end{figure}
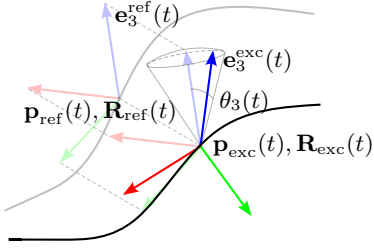

The execution of skills is guaranteed to satisfy the time-varying conic constraints in Fig.~\ref{fig:CCBF_illustration} by solving 
\eqref{eq:ccqp}.
%
In Fig. \ref{fig:overview}, two distinct \acp{ds} run simultaneously: (\emph{i}) the executor \ac{ds}, which generates control input for the actively executed rotational trajectory $\bm{R}_\mathrm{exc}(t)$, and (\emph{ii}) the reference \ac{ds}, which provides the reference rotational trajectory $\bm{R}_\mathrm{ref}(t)$ for the time-varying conic constraints. Their respective axes are denoted as $\bm{e}_i^\mathrm{exc} = \bm{R}_\mathrm{exc}\bm{e}_i^{\mathrm{w}}$ and $\bm{e}_i^\mathrm{ref} = \bm{R}_\mathrm{ref}\bm{e}_i^{\mathrm{w}}$. At each time step, we compute angular velocities $\bm{\omega}_{\mathrm{ref}}$ and $\bm{\omega}_{\mathrm{exc}}$ using  \eqref{eq:so3ds} and determine the cone angle $\bm{\theta}$ with the learned \ac{lwr} model $\bm{f}_\theta$. The summation $\bm{u}_0 = \bm{\omega}_{\mathrm{exc}} + \bm{u}_h + \bm{u}_{p}$ serves as the reference control input to the \ac{qp}, where $\bm{u}_h$ and $\bm{u}_p$ denote human inputs and perturbations. The reference trajectory follows exactly the learned \ac{ds} $\dot{\bm{R}}_\mathrm{ref} =\bm{R}_\mathrm{ref}[\bm{\omega}
_\mathrm{ref}]_{\times}$, while the execution follows $\dot{\bm{R}}_\mathrm{exc} =\bm{R}_\mathrm{exc}[\bm{u}^*]_{\times}$, where $\bm{u}^*$ is the solution to the following \ac{qp}:
%
\begin{equation}
\label{eq:ccqp}
\begin{aligned} 
&\underset{\bm{u} \in \mathcal{U}}{\operatorname{argmin}}\; \frac{1}{2}\|\bm{u}-\bm{u}_0(\bm{R}_\mathrm{exc}, t)\|_2^2 \hspace*{2.7cm}(\text{CCBF-QP})\\ &\text{ s.t. } \dot{h}_i(\bm{R}_\mathrm{exc}, t)\geq-\alpha\left(h_i(\bm{R}_\mathrm{exc}, t)\right), i \in\{1,2,3\}.
\end{aligned}
\end{equation}
The \ac{ccbf} condition explicitly writes as
\begin{equation*}
    \begin{split}
        -\bm{e}_i^\mathrm{ref}\trsp &  \bm{R}_\mathrm{exc}\left[\bm{e}_i^{\mathrm{w}}\right]_{\times} \bm{u} - \bm{e}_i^\mathrm{exc}\trsp \bm{R}_\mathrm{ref}\left[\bm{e}_i^{\mathrm{w}}\right]_{\times} \bm{\omega}_{\mathrm{ref}} \\
        &+\frac{d}{dt}\operatorname{cos}(\theta_i)+\alpha\left(\bm{e}_i^{\mathrm{ref}}\trsp\bm{e}_i^{\mathrm{exc}} - \operatorname{cos}(\theta_i)\right) \geq 0.
    \end{split}
\end{equation*}

Notably, the executor \ac{ds} can be arbitrarily selected since it will remain confined within the time-varying cones. For simplicity and more accurate reference trajectory tracking, we opt for the same executor \ac{ds} as the reference \ac{ds}. However, the reference \ac{ds} evolving independently of human input could potentially pose challenges in assisted teleoperation, as the human operator may lose control authority over the execution. To address this, a human operator can modulate the actual velocity using an additional continuous input device, such as a foot pedal.



\section{Experimental Results}

\subsection{Simulation}
\begin{figure}[t!]
\centering
    \def\svgwidth{\columnwidth}
    {\fontsize{7}{7}
    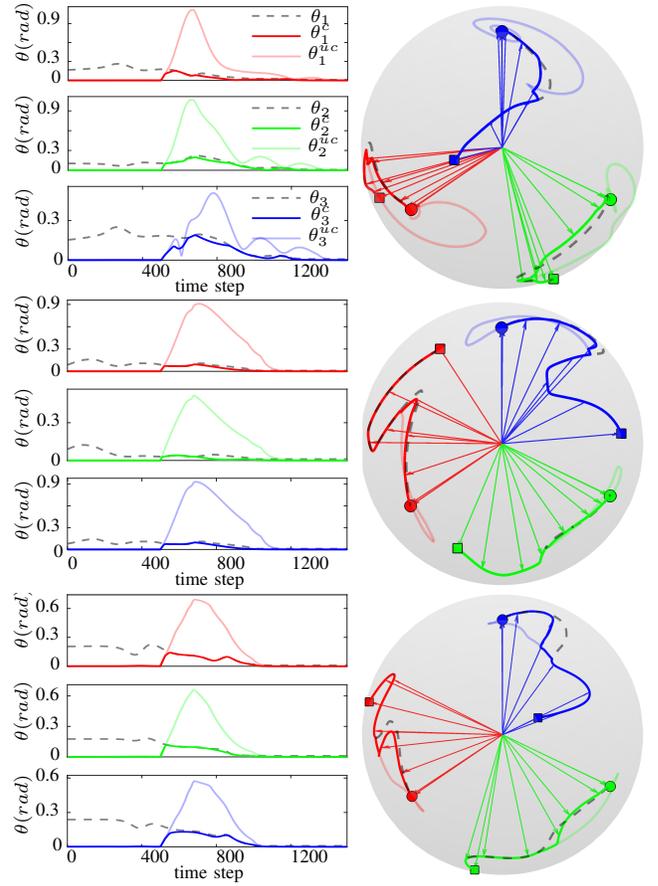}
    

     \caption{Temporal evolution of cone angles and orientation trajectories for the L/N/W shapes (from top to bottom). Orientation is depicted in transformed frames, with square and circle markers denoting initial orientations and goals. On the left, deep/light solid lines represent the angles between constrained/unconstrained axes and reference axes, denoted as $\theta_i^c$ and $\theta_i^{uc}$ respectively. Red, green, and blue correspond to the $x$-axis, $y$-axis, and $z$-axis. Dashed black lines indicate the learned cone angles. On the right, the trajectories in solid deep/light blue correspond to constrained/unconstrained executions, and the dashed black trajectories represent the reference trajectories.}
\label{fig:simulation_results}
\end{figure}
In simulation, we used orientation trajectories from \ac{rlasa}~\cite{saveriano2023learning}, an extension to LASA handwriting dataset~\cite{khansari2011learning}. We empirically set 5 Gaussian components for learning the translational and 8 for the rotational \acp{ds}. For \ac{lwr} of cone angles, we opted for a polynomial degree of 5 and employed 10 components, which appeared to offer well-balanced fitting accuracy. To avoid an empty admissible set, we artificially set the minimum value of the cone angle to $0.01\,$rad. 
During execution, we employed a time step of $0.003\,$s, starting at the average of the demonstrations and with the goal set as the identity matrix. Approximately $1.5\,$s after the start, we introduced a smooth step function lasting $0.5\,$s as the disturbance. 

Fig. \ref{fig:simulation_results} illustrates the disturbed execution of the L/N/W shapes from the \ac{rlasa}. Firstly, we noticed that both trajectories with and without the \ac{ccbfqp} in all cases converged to the identity matrix, which verified the asymptotic stability of the \ac{ds} learned with \ac{physicsgmm}. However, all the unconstrained orientation trajectories deviated significantly from the reference trajectories after being disturbed, partly due to the inherent convergence behavior. In contrast, the constrained trajectories remained within the region defined by the time-varying cones, closely tracking the reference trajectories. To more clearly showcase the satisfaction of constraints, we calculated the \ac{nacv} for all axes, i.e.
\begin{equation*}
    \mathrm{NACV} = \frac{1}{3T}\Sigma_{t=0}^{T}\Sigma_{i=1}^{3}\operatorname{max}\{0, (\theta_{i}^{uc/c}(t) - \theta_{i}(t))/ \theta_{i}(t)\},
\end{equation*} 
where $\theta_i^{uc/c}$ refers to the angles between the actual and the reference cone axes for unconstrained/constrained runs. We obtained $\mathrm{NACV} = 2.033 \pm 0.844$ for unconstrained executions, and, as expected, $\mathrm{NACV} = 0$ (no violations) for executions subject to constraints.

\begin{figure}[t!]
\centering
    \includegraphics[width=\columnwidth]{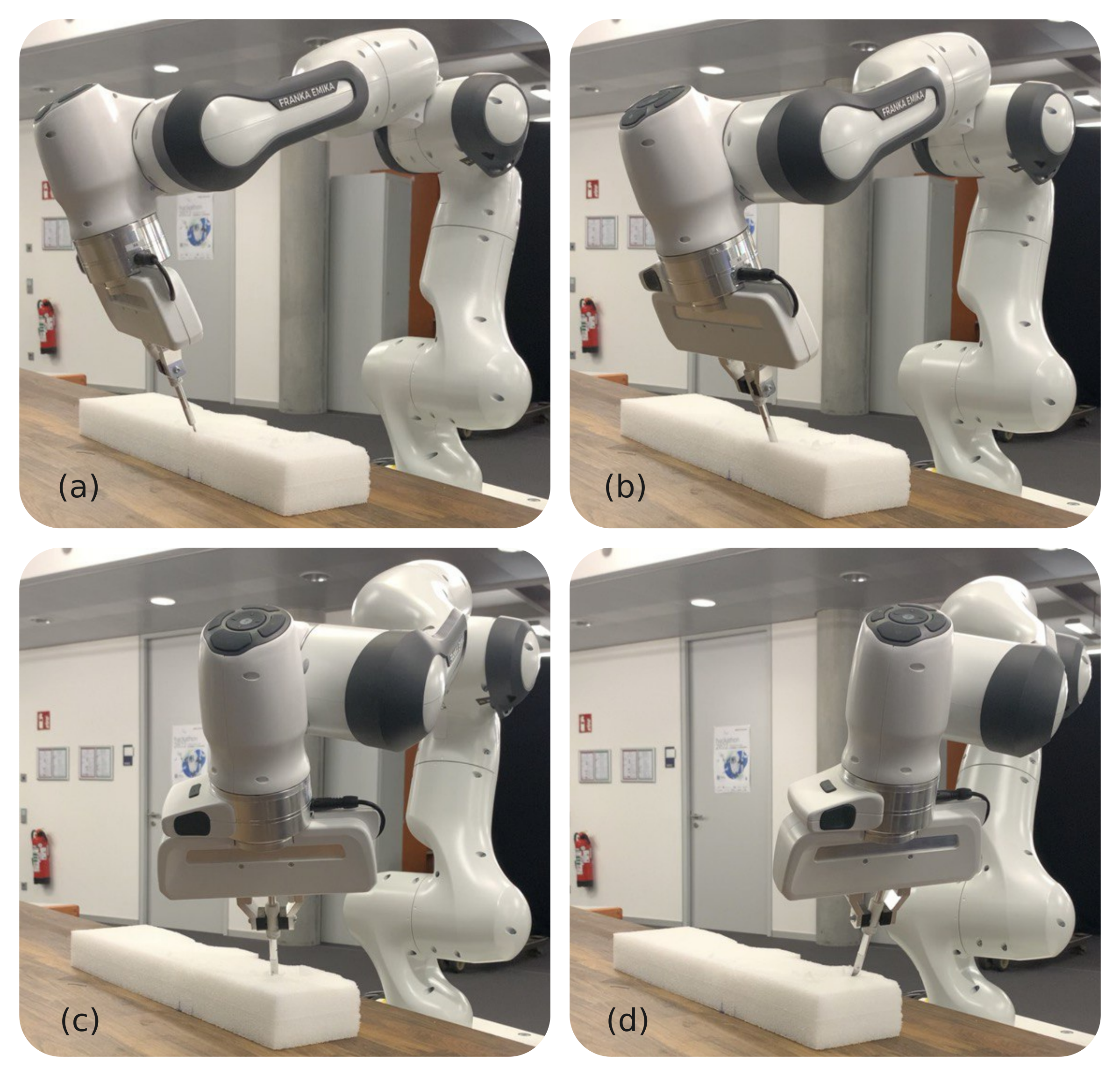}
\vspace{-.5cm}
\caption{Execution of the cutting skill in assisted teleoperation. }
\label{fig:snapshots}
\end{figure}
\subsection{Robot Experiment}

In this experiment, we focused on learning and executing a cutting skill in an assisted teleoperation scenario. Cutting tasks demand continuous orientation adjustments along the trajectory with particular significance placed on the tilt angle of the blades, since it determines incision shapes as illustrated in Fig. \ref{fig:cutting setup}.  In applications such as robotic surgery for tissue resection, operators may require precise control over the incision shape by adjusting the blade tilt during skill execution. Meanwhile, it's crucial to confine the tilt angle within a predefined range to ensure safety. Thus, we imposed constraints on the tilt angle of the blade, which coincides with the $z$-axis of the \ac{ee} frame.

Our hardware setup involved a 7 degrees of freedom manipulator \emph{Franka Research 3} for skill execution and a haptic device \emph{lambda.7} for teleoperation. The manipulator was controlled by a Cartesian impedance controller with a translational stiffness matrix of $\mathrm{diag}([1500, 1500, 1500])\,$N/m and a rotational stiffness matrix of $\mathrm{diag}([75, 75, 75])\,$Nm/rad. The damping matrices were set for critical damping.

To capture the cutting skill, We recorded five kinesthetic demonstrations. The hyperparameters of \ac{gmm} and \ac{lwr} remained the same as in the simulation. The minimum threshold of cone angle was set to $0.02\,$rad.

    

\begin{figure}[t!]
\centering
\def\svgwidth{\columnwidth}
    {\fontsize{7}{7}
    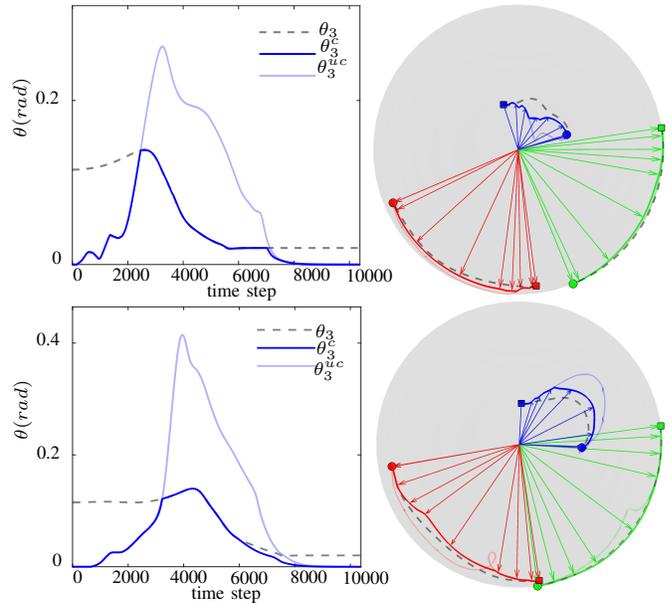}

 \caption{Cone angles over time and rotational trajectories of the \ac{ee} frame of the robot. On the left, solid lines in deep/light blue represent the angles between constrained/unconstrained z-axes and reference z-axes, denoted as $\theta_3^c$ and $\theta_3^{uc}$ respectively, under human intervention. The black dashed lines indicate the learned cone angle bounds. On the right, the trajectories in solid deep/light blue correspond to reproductions w/o \ac{ccbfqp}, while the dashed black trajectories represent the reference trajectories.}
\label{fig:experiment_result}
\end{figure}

In the performance evaluation, we invited six users, each performing the test twice. The users introduced different inclinations of the cutting edge, potentially or deliberately exceeding the learned conic constraints around the $z$-axis of the \ac{ee} frame. Fig. \ref{fig:experiment_result} shows the temporal evolution of cone angles and the rotational trajectories of the \ac{ee} frame in two independent runs. We observed that the executions with \ac{cbf} consistently fulfilled the learned constraints (the \ac{nacv} of the $z$-axis is $\mathrm{NACV_z} = 0$), while unconstrained ones deviated far from the reference trajectories under excessive operator inputs ($\mathrm{NACV_z} =  2.218 \pm 1.025$). 

\section{CONCLUSIONS}
In summary, we presented an innovative method for the safe execution of learned rotation skills by leveraging the combination of (\emph{i}) a stable \ac{ds} on $SO(3)$ and  (\emph{ii}) \acp{ccbf} that guarantee satisfaction of (\emph{iii}) time-varying conic constraints extracted from the variability of demonstrations. 
We showed in simulation that our approach ensured that the learned \ac{ds} converged to the goal, evolved closely to the reference trajectory, and always stayed within the region defined by the conic constraints.
Furthermore, we have showcased the practical applicability of our approach in an assisted teleoperation scenario for cutting skills. In our future work, we intend to enhance the constraint learning method, currently restricted to trajectories whose distance to the target monotonically decreases. Moreover, we aim to improve the approach's usability by encompassing both translational and rotational motions based on coupled \acp{ds}~\cite{shukla2012coupled}.








\bibliographystyle{IEEEtran}
\bibliography{ref}

\begin{thebibliography}{10}
\providecommand{\url}[1]{#1}
\csname url@rmstyle\endcsname
\providecommand{\newblock}{\relax}
\providecommand{\bibinfo}[2]{#2}
\providecommand\BIBentrySTDinterwordspacing{\spaceskip=0pt\relax}
\providecommand\BIBentryALTinterwordstretchfactor{4}
\providecommand\BIBentryALTinterwordspacing{\spaceskip=\fontdimen2\font plus
\BIBentryALTinterwordstretchfactor\fontdimen3\font minus
  \fontdimen4\font\relax}
\providecommand\BIBforeignlanguage[2]{{%
\expandafter\ifx\csname l@#1\endcsname\relax
\typeout{** WARNING: IEEEtran.bst: No hyphenation pattern has been}%
\typeout{** loaded for the language `#1'. Using the pattern for}%
\typeout{** the default language instead.}%
\else
\language=\csname l@#1\endcsname
\fi
#2}}

\bibitem{billard2008survey}
A.~Billard, S.~Calinon, R.~Dillmann, and S.~Schaal, ``Survey: Robot programming
  by demonstration,'' Springrer, Tech. Rep., 2008.

\bibitem{khansari2011learning}
S.~M. Khansari-Zadeh and A.~Billard, ``Learning stable nonlinear dynamical
  systems with gaussian mixture models,'' \emph{IEEE Transactions on Robotics},
  vol.~27, no.~5, pp. 943--957, 2011.

\bibitem{figueroa2022locally}
N.~Figueroa and A.~Billard, ``Locally active globally stable dynamical systems:
  Theory, learning, and experiments,'' \emph{The International Journal of
  Robotics Research}, vol.~41, no.~3, pp. 312--347, 2022.

\bibitem{figueroa2018physically}
N.~B. Figueroa~Fernandez and A.~Billard, ``A physically-consistent bayesian
  non-parametric mixture model for dynamical system learning,''
  \emph{Proceedings of Machine Learning Research}, 2018.

\bibitem{ibuki2020distributed}
T.~Ibuki, S.~Wilson, A.~D. Ames, and M.~Egerstedt, ``Distributed collision-free
  motion coordination on a sphere: A conic control barrier function approach,''
  \emph{IEEE Control Systems Letters}, vol.~4, no.~4, pp. 976--981, 2020.

\bibitem{ijspeert2013dynamical}
A.~J. Ijspeert, J.~Nakanishi, H.~Hoffmann, P.~Pastor, and S.~Schaal,
  ``Dynamical movement primitives: learning attractor models for motor
  behaviors,'' \emph{Neural computation}, vol.~25, no.~2, pp. 328--373, 2013.

\bibitem{saveriano2021dynamic}
M.~Saveriano, F.~J. Abu-Dakka, A.~Kramberger, and L.~Peternel, ``Dynamic
  movement primitives in robotics: A tutorial survey,'' \emph{The International
  Journal of Robotics Research}, 2023.

\bibitem{zeestraten2017approach}
M.~J. Zeestraten, I.~Havoutis, J.~Silv{\'e}rio, S.~Calinon, and D.~G. Caldwell,
  ``An approach for imitation learning on riemannian manifolds,'' \emph{IEEE
  Robotics and Automation Letters}, vol.~2, no.~3, pp. 1240--1247, 2017.

\bibitem{abu2022unified}
F.~J. Abu-Dakka, M.~Saveriano, and V.~Kyrki, ``A unified formulation of
  geometry-aware dynamic movement primitives,'' \emph{arXiv preprint
  arXiv:2203.03374}, 2022.

\bibitem{saveriano2023learning}
M.~Saveriano, F.~J. Abu-Dakka, and V.~Kyrki, ``Learning stable robotic skills
  on riemannian manifolds,'' \emph{Robotics and Autonomous Systems}, p. 104510,
  2023.

\bibitem{ureche2015task}
A.~L.~P. Ureche, K.~Umezawa, Y.~Nakamura, and A.~Billard, ``Task
  parameterization using continuous constraints extracted from human
  demonstrations,'' \emph{IEEE Transactions on Robotics}, vol.~31, no.~6, pp.
  1458--1471, 2015.

\bibitem{menner2019constrained}
M.~Menner, P.~Worsnop, and M.~N. Zeilinger, ``Constrained inverse optimal
  control with application to a human manipulation task,'' \emph{IEEE
  Transactions on Control Systems Technology}, vol.~29, no.~2, pp. 826--834,
  2019.

\bibitem{saveriano2019learning}
M.~Saveriano and D.~Lee, ``Learning barrier functions for constrained motion
  planning with dynamical systems,'' in \emph{2019 IEEE/RSJ International
  Conference on Intelligent Robots and Systems (IROS)}.\hskip 1em plus 0.5em
  minus 0.4em\relax IEEE, 2019, pp. 112--119.

\bibitem{davoodi2021probabilistic}
M.~Davoodi, A.~Iqbal, J.~M. Cloud, W.~J. Beksi, and N.~R. Gans, ``Probabilistic
  movement primitive control via control barrier functions,'' in \emph{2021
  IEEE 17th International Conference on Automation Science and Engineering
  (CASE)}.\hskip 1em plus 0.5em minus 0.4em\relax IEEE, 2021, pp. 697--703.

\bibitem{sutanto2021learning}
G.~Sutanto, I.~R. Fern{\'a}ndez, P.~Englert, R.~K. Ramachandran, and
  G.~Sukhatme, ``Learning equality constraints for motion planning on
  manifolds,'' in \emph{Conference on Robot Learning}.\hskip 1em plus 0.5em
  minus 0.4em\relax PMLR, 2021, pp. 2292--2305.

\bibitem{mehr2016inferring}
N.~Mehr, R.~Horowitz, and A.~D. Dragan, ``Inferring and assisting with
  constraints in shared autonomy,'' in \emph{2016 IEEE 55th Conference on
  Decision and Control (CDC)}.\hskip 1em plus 0.5em minus 0.4em\relax IEEE,
  2016, pp. 6689--6696.

\bibitem{perez2017c}
C.~P{\'e}rez-D'Arpino and J.~A. Shah, ``C-learn: Learning geometric constraints
  from demonstrations for multi-step manipulation in shared autonomy,'' in
  \emph{2017 IEEE International Conference on Robotics and Automation
  (ICRA)}.\hskip 1em plus 0.5em minus 0.4em\relax IEEE, 2017, pp. 4058--4065.

\bibitem{chou2021learning}
G.~Chou, D.~Berenson, and N.~Ozay, ``Learning constraints from demonstrations
  with grid and parametric representations,'' \emph{The International Journal
  of Robotics Research}, vol.~40, no. 10-11, pp. 1255--1283, 2021.

\bibitem{ames2019control}
A.~D. Ames, S.~Coogan, M.~Egerstedt, G.~Notomista, K.~Sreenath, and P.~Tabuada,
  ``Control barrier functions: Theory and applications,'' in \emph{2019 18th
  European control conference (ECC)}.\hskip 1em plus 0.5em minus 0.4em\relax
  IEEE, 2019, pp. 3420--3431.

\bibitem{IGARASHI2019735}
M.~Igarashi, I.~Tezuka, and H.~Nakamura, ``Time-varying control barrier
  function and its application to environment-adaptive human assist control,''
  \emph{IFAC Symposium on Nonlinear Control Systems}, vol.~52, no.~16, pp.
  735--740, 2019.

\bibitem{wu2016safety}
G.~Wu and K.~Sreenath, ``Safety-critical geometric control for systems on
  manifolds subject to time-varying constraints,'' \emph{IEEE Transactions on
  Automatic Control (TAC), in review}, 2016.

\bibitem{tan2020construction}
X.~Tan and D.~V. Dimarogonas, ``Construction of control barrier function and
  $\mathcal{C}^2$ reference trajectory for constrained attitude maneuvers,'' in
  \emph{2020 59th IEEE Conference on Decision and Control (CDC)}.\hskip 1em
  plus 0.5em minus 0.4em\relax IEEE, 2020, pp. 3329--3334.

\bibitem{rosenberg1993virtual}
L.~B. Rosenberg, ``Virtual fixtures: Perceptual tools for telerobotic
  manipulation,'' in \emph{Proceedings of IEEE virtual reality annual
  international symposium}.\hskip 1em plus 0.5em minus 0.4em\relax Ieee, 1993,
  pp. 76--82.

\bibitem{ewerton2020assisted}
M.~Ewerton, O.~Arenz, and J.~Peters, ``Assisted teleoperation in changing
  environments with a mixture of virtual guides,'' \emph{Advanced Robotics},
  vol.~34, no.~18, pp. 1157--1170, 2020.

\bibitem{dragan2013policy}
A.~D. Dragan and S.~S. Srinivasa, ``A policy-blending formalism for shared
  control,'' \emph{The International Journal of Robotics Research}, vol.~32,
  no.~7, pp. 790--805, 2013.

\bibitem{javdani2018shared}
S.~Javdani, H.~Admoni, S.~Pellegrinelli, S.~S. Srinivasa, and J.~A. Bagnell,
  ``Shared autonomy via hindsight optimization for teleoperation and teaming,''
  \emph{The International Journal of Robotics Research}, vol.~37, no.~7, pp.
  717--742, 2018.

\bibitem{havoutis2016learning}
I.~Havoutis and S.~Calinon, ``Learning assistive teleoperation behaviors from
  demonstration,'' in \emph{2016 IEEE International Symposium on Safety,
  Security, and Rescue Robotics (SSRR)}.\hskip 1em plus 0.5em minus 0.4em\relax
  IEEE, 2016, pp. 258--263.

\bibitem{havoutis2017supervisory}
------, ``Supervisory teleoperation with online learning and optimal control,''
  in \emph{2017 IEEE International Conference on Robotics and Automation
  (ICRA)}.\hskip 1em plus 0.5em minus 0.4em\relax IEEE, 2017, pp. 1534--1540.

\bibitem{mower2021skill}
C.~E. Mower, J.~Moura, and S.~Vijayakumar, ``Skill-based shared control.'' in
  \emph{Robotics: Science and Systems}, 2021, pp. 1--10.

\bibitem{bullo2019geometric}
F.~Bullo and A.~D. Lewis, \emph{Geometric control of mechanical systems:
  modeling, analysis, and design for simple mechanical control systems}.\hskip
  1em plus 0.5em minus 0.4em\relax Springer, 2019, vol.~49.

\bibitem{lindemann2018control}
L.~Lindemann and D.~V. Dimarogonas, ``Control barrier functions for signal
  temporal logic tasks,'' \emph{IEEE control systems letters}, vol.~3, no.~1,
  pp. 96--101, 2018.

\bibitem{murray2017mathematical}
R.~M. Murray, Z.~Li, and S.~S. Sastry, \emph{A mathematical introduction to
  robotic manipulation}.\hskip 1em plus 0.5em minus 0.4em\relax CRC press,
  2017.

\bibitem{Berkane2017Construction}
S.~Berkane and A.~Tayebi, ``Construction of synergistic potential functions on
  so(3) with application to velocity-free hybrid attitude stabilization,''
  \emph{IEEE Transactions on Automatic Control}, vol.~62, no.~1, pp. 495--501,
  2017.

\bibitem{shukla2012coupled}
A.~Shukla and A.~Billard, ``Coupled dynamical system based arm--hand grasping
  model for learning fast adaptation strategies,'' \emph{Robotics and
  Autonomous Systems}, vol.~60, no.~3, pp. 424--440, 2012.

\end{thebibliography}
\end{document}